\documentclass[journal=jacsat,manuscript=article]{achemso}

\usepackage[version=3]{mhchem} 
\usepackage{amsmath,amsfonts,amssymb}
\usepackage{tabularx}
\usepackage{booktabs} 
\usepackage{adjustbox}
\usepackage{hyperref}
\usepackage{siunitx}
\DeclareSIUnit\angstrom{\text {Å}}

\newcommand{\R}{\mathbb{R}}
\author{Zhonglin Cao}
\affiliation[meche]
{Department of Mechanical Engineering, Carnegie Mellon University, Pittsburgh PA, USA 15213}
\altaffiliation{Joint First Authorship}
\author{Rishikesh Magar}
\affiliation[meche]
{Department of Mechanical Engineering, Carnegie Mellon University, Pittsburgh PA, USA 15213}
\altaffiliation{Joint First Authorship}
\author{Yuyang Wang}
\affiliation[meche]
{Department of Mechanical Engineering, Carnegie Mellon University, Pittsburgh PA, USA 15213}
\author{Amir Barati Farimani}
\email{barati@cmu.edu}
\affiliation[meche]
{Department of Mechanical Engineering, Carnegie Mellon University, Pittsburgh PA, USA 15213}
\alsoaffiliation[biomed]
{Department of Chemical Engineering, Carnegie Mellon University, Pittsburgh PA, USA 15213}
\alsoaffiliation[mld]
{Machine Learning Department, Carnegie Mellon University, Pittsburgh PA, USA 15213}

\title{MOFormer: Self-Supervised Transformer model for Metal-Organic Framework Property Prediction}
\abbreviations{IR,NMR,UV}

\begin{document}







\begin{abstract}
Metal-Organic Frameworks (MOFs) are materials with a high degree of porosity that can be used for applications in energy storage, water desalination, gas storage, and gas separation. However, the chemical space of MOFs is close to an infinite size due to the large variety of possible combinations of building blocks and topology. Discovering the optimal MOFs for specific applications requires an efficient and accurate search over an enormous number of potential candidates. Previous high-throughput screening methods using computational simulations like DFT can be time-consuming. Such methods also require optimizing 3D atomic structure of MOFs, which adds one extra step when evaluating hypothetical MOFs. In this work, we propose a structure-agnostic deep learning method based on the Transformer model, named as MOFormer, for property predictions of MOFs. The MOFormer takes a text string representation of MOF (MOFid) as input, thus circumventing the need of obtaining the 3D structure of hypothetical MOF and accelerating the screening process. Furthermore, we introduce a self-supervised learning framework that pretrains the MOFormer via maximizing the cross-correlation between its structure-agnostic representations and structure-based representations of crystal graph convolutional neural network (CGCNN) on $>$400k publicly available MOF data. Using self-supervised learning allows the MOFormer to intrinsically learn 3D structural information though it is not included in the input. Experiments show that pretraining improved the prediction accuracy of both models on various downstream prediction tasks. Furthermore, we revealed that MOFormer can be more data-efficient on quantum-chemical property prediction than structure-based CGCNN when training data is limited. Overall, MOFormer provides a novel perspective on efficient MOF design using deep learning.

\end{abstract}

\section{Introduction}
Metal-organic frameworks (MOFs) are a type of porous crystalline materials\cite{james2003metal, zhou2012introduction}, which have been extensively researched during the past several decades. Research interests have been induced by the porous structure and versatile nature of MOFs on their potential applications such as gas adsorption\cite{ahmed2019exceptional, boyd2019data, wilmer2012large}, water harvesting and desalination\cite{almassad2022environmentally, hanikel2020mof, cao2019water}, and energy storage\cite{baumann2019metal,zhao2016metal,xu2017exploring}. MOFs typically consist of several building blocks, including metal-based metal nodes and organic linkers \cite{sharp2021nanoconfinement, moosavi2020understanding, boyd2019data}. The assembly of those building blocks following certain topologies generates the 2-dimensional or 3-dimensional porous structures of MOFs. Because of the countless possible combinations of metal nodes, organic linkers, and topologies\cite{moosavi2020understanding, falcaro2011new}, there is a sheer amount of MOFs with different mechanical properties and surface chemistry. Given the enormous variety of possible MOF structures, rapidly and inexpensively selecting the potential top performers for each specific task can be challenging. High-throughput screening with computational tools such as molecular simulation \cite{wilmer2012large, ozcan2020modeling} or density functional theory (DFT) \cite{canepa2013high, nandy2021computational} has been widely used to evaluate the properties of MOFs. Without the need to experimentally synthesize MOF structures, those computational tools accelerate the screening process and allow researchers to screen hundreds of thousands of hypothetical MOF structures for their performance in different applications.

Recently, machine learning (ML) models have become increasingly popular in the field of MOF property prediction\cite{fung2021benchmarking, burner2020high, altintas2021machine, lee2021computational, choudhary2022graph, moghadam2019structure, nandy2021using}. The advantage of the ML models over the simulation methods is their instantaneous inference of the properties of MOFs. In contrast, the simulation methods require a computationally expensive rerun for every new MOF. In the last decade, multiple large scale MOF dataset are released, including the CoRE MOF 2019\cite{chung2019advances}, hypothetical MOFs\cite{wilmer2012large}, and QMOF\cite{rosen2021machine}. These datasets contain the atomic structures of MOFs and their properties like \ce{CO_2} absorption and band gap. These datasets are large enough to train accurate data-driven ML models for the prediction of MOF properties. Handcrafted geometrical features such as large cavity diameter and pore limiting diameter have been used as input to a multilayer perceptron (MLP) to predict MOFs properties \cite{burner2020high, moghadam2019structure}. Although the training of MLP with a few layers can be fast, this method suffers from underwhelming accuracy due to the simplicity of network architecture. Moreover, selecting features requires extensive domain knowledge from the researchers and optimized 3D structures of MOFs, thus making this method less generic. Given the aforementioned drawbacks, a novel method that can achieve high accuracy with a more generic input of MOFs representations should be pursued. Wang et al. \cite{wang2020accelerating} utilize the crystal graph convolutional neural network (CGCNN) \cite{xie2018crystal} to predict methane absorption of MOFs. CGCNN is a prevalent model with has an architecture designed specifically for crystalline materials. It takes the type and the 3D coordinates of atoms in the crystalline materials as input and constructs a crystal graph. CGCNN can extract features that encode rich chemical information through convolution operations on the crystal graph. However, one drawback of using CGCNN for MOFs property prediction is that it requires the optimized 3D atomic structures of MOFs which are computationally expensive to obtain. In addition, some large MOF structures consist of hundred or even thousand of atoms, thus rendering crystal graph for them can be memory-inefficient.

Enlightened by the fact that all MOFs are combinations of metal nodes, organic linkers, and topologies, Bucior et al.\cite{bucior2019identification} proposed a text string representation of MOFs called MOFid. The two core sections of a typical MOFid include the chemical information of building blocks in the format of SMILES\cite{weininger1988smiles} and the topology and the catenation of the MOF structure. The building blocks are represented by an extensively used string representation of molecules called SMILES \cite{weininger1988smiles}. The topology and catenation are each represented by a code adopted from the Reticular Chemistry Structure Resource (RCSR) database\cite{o2008reticular}. Therefore, MOFid is a concise text string representation of MOFs that preserves the chemical and the majority of the structural information through topology encoding. The MOFid text based representation enables the application of language ML models that take text string as input for MOF property prediction.

In this work, we proposed and developed a Transformer-based language model for MOF property prediction. Transformer and its variants have become the top choice for the natural language processing tasks since publication in 2017 by Vaswani et al\cite{vaswani2017attention}. The multi-head attention mechanism allows the Transformer model to learn contextual information in a sequence without suffering from long-range dependency\cite{bahdanau2014neural, hochreiter1997long}. With its success in processing long sequential data, Transformer and its other variants are also adopted for chemistry or bioinformatics application such as molecular\cite{schwaller2019molecular, schwaller2021mapping, xu2022transpolymer} and protein\cite{elnaggar2021prottrans} property prediction. The Transformer model in our work, named as MOFormer, takes a modified MOFid as input to make predictions of various MOF properties. The advantage of this method is that it does not require the 3D atomic structure of the MOF (structure-agnostic), thus enabling a much faster and more flexible exploration of the hypothetical MOF space. In practice, pretraining the Transformer model in a self-supervised manner\cite{devlin2018bert, liu2019roberta, elnaggar2021prottrans, wang2022molecular, wang2022improving} leverages large quantity of unlabeled data to help the MOFormer learn a more robust representation of the sequence, and further improve its performance in downstream tasks. To take advantage of pre-training benefits, we also added a self-supervised learning framework in which the MOFormer and the CGCNN model are jointly pretrained with $>$400k MOF structures. Dimensionality reduction tools are used to visualize the latent representation learned by both models to provide insight into their performance characteristics. Visualization of attention weights in MOFormer demonstrates that MOFormer learns MOF representations based on some key atoms and topology. Lastly, we compared the data efficiency of models to show which one is a better choice when training data is limited. 

\section{Methods}
\subsection{MOFid tokenization and Transformer}
\begin{figure}[htb!]
    \centering
    \includegraphics[width=\linewidth]{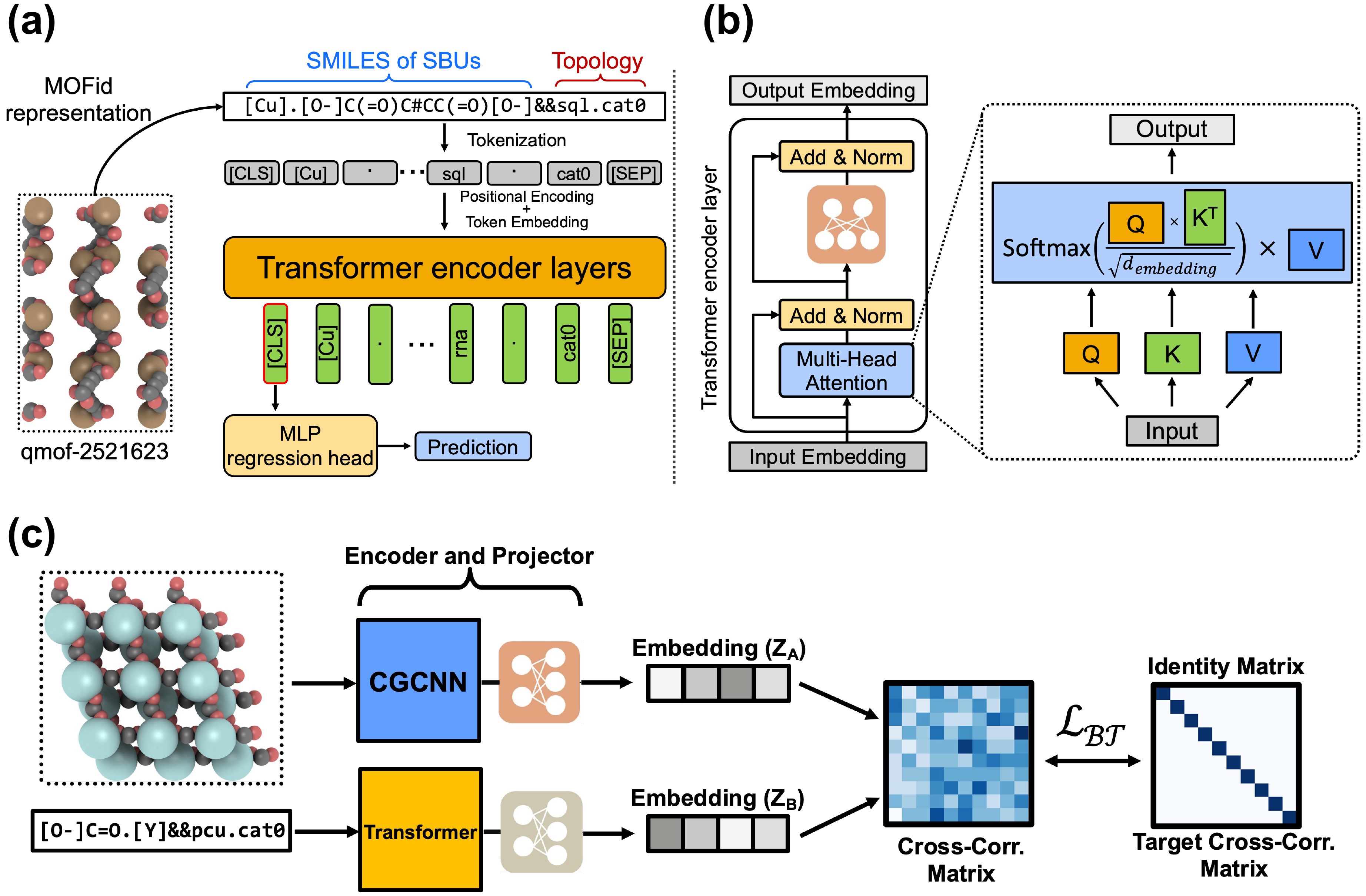}
    \caption{\textbf{(a)} The pipeline of the MOFormer model. A MOFid of a MOF (qmof-2521623 is used as an example) is the input to the model. The MOFid is converted into tokenized sequence before embedded and applied with positional encoding. The sequence is then fed into multiple Transformer encoder layers. The learned embedding of the first token will be used as input to a MLP regression head for downstream prediction task. \textbf{(b)} A schematic showing the details of each Transformer encoder layer. Embeddings of the sequence pass through the multi-head scaled dot-product attention layer and then a MLP. Residue connection and layer normalization is adopted for both the attention and the MLP. \textbf{(c)} The self-supervised learning framework with CGCNN and MOFormer. The 3D structure and the MOFid of the same MOF are fed into the CGCNN and MOFormer, respectively for representation learning. The MLP head following each models project the representations into embeddings ($Z_A$ and $Z_B$). A cross correlation matrix is then constructed using the embeddings. Barlow Twins loss is applied to optimize the cross correlation matrix to be as close as possible to an identity matrix.}
    \label{fig:model}
\end{figure}

The MOFormer is built upon the encoder part of the Transformer model that takes tokenized MOFid as input (Figure.~\ref{fig:model}a). The MOFid tokenizer is a customized version of the SMILES tokenizer\cite{schwaller2018found}. The SMILES strings of all secondary building units (SBUs) of the MOFid is tokenized by the SMILES tokenizer, while the topology and catenation section of the MOFid is separately tokenized based on the topology encoding adopted from RCSR\cite{o2008reticular}. The tokens from both sections are then connected by a separator token ``\texttt{\&\&}". The tokenization process followed the BERT\cite{devlin2018bert} to add a \texttt{[CLS]} token and a \texttt{[SEP]} token at the beginning and the end of the sequence to symbolize the start and the end, respectively. Since the tokenized sequences conform to a fixed length of 512, sequences longer than the fixed length are truncated and the sequences shorter than that are padded with special tokens \texttt{[PAD]}. 

Tokenized sequence is embedded and combined with a positional encoding (Figure.~\ref{fig:model}a) to include information about the relative and absolute position of each token. The position encoding is calculated by:
\begin{equation}
\begin{split}
    \text{PE}_{(pos, 2i)} &= \text{sin}(\frac{pos}{10000^{2i/d_{emb}}})\\
    \text{PE}_{(pos, 2i+1)} &= \text{cos}(\frac{pos}{10000^{2i/d_{emb}}})
\end{split}
\end{equation}
where $pos$ is the position of the token in the sequence, $i$ is the index of dimension, and $d_{emb}=512$ is the embedding dimension. The Transformer model is a deep neural network model built upon the self-attention mechanism (detailed in the supporting information). Each of the Transformer encoder layer consists of a multi-head attention layer followed by a simple feed-forward multilayer perceptron (MLP). Residue connection\cite{he2016deep} and layer normalization\cite{ba2016layer} is adopted for both the attention and the MLP. In each head of the attention layer (Figure.~\ref{fig:model}b), the input sequence embedding $X$ is multiplied with three learnable weight vectors $W_q$, $W_k$, and $W_v$ to be converted to the query, key, and value vector ($Q$, $K$, $V$). The scaled dot-product attention $A$ is then calculated by the equation:
\begin{equation}
    A = \text{softmax}(\frac{QK^\top}{\sqrt{d_{k}}})V
\end{equation}
where $d_k$ is the dimension of $Q$ and $K$ (detailed in the supporting information). The randomly initialized $W_q$, $W_k$, and $W_v$ vectors in each head allow the model to learn the contextual information between tokens in different representation subspaces\cite{vaswani2017attention}. Attention from all heads are concatenated together and then fed into the MLP for the projected output embedding, which is of the same size of the input embedding. Given that the self-mechanism can incorporate the information of the whole sequence into each one of the token embeddings, theoretically any one of the embeddings can be used as a representation of the whole sequence. Therefore, we followed the common practice of related works\cite{schwaller2019molecular, schwaller2021extraction, schwaller2021mapping, dosovitskiy2020image} to use the embedding of the first token, \texttt{[CLS]}, for further supervised learning tasks.

\subsection{Self-supervised pretraining with CGCNN}
We introduce a self-supervised learning (SSL) paradigm for MOF representation learning. We designed the framework by taking into consideration two modalities of data including the text and graph information. One of the modalities is the text string representation (MOFid) that encapsulates building blocks' stoichiometry and bonds (SMILES) and the topology of the MOF. The text string information is processed by the MOFormer. One of the limitations of text string data is the lack of availability of information about the geometry and the neighborhood of atoms creating an information bottleneck for the tex string input based models. The structure agnostic nature of the text string input can prevent the transformer to achieve higher performance than the graph based models. To mitigate these issues of the MOFormer framework, we introduce SSL pretraining with CGCNN. We leverage structure as a modality with CGCNN\cite{xie2018crystal}. The CGCNN model takes as input the structure of the MOF. The structure data consists of the atom information along with the neighborhood information which is critical in property prediction tasks. To implement the SSL pipeline, we take inspiration from Crystal Twins (CT) framework\cite{magar2022crystal}. The CT model makes use of the Barlow Twins loss function introduced by Zbontar et al\cite{zbontar2021barlow} and SimSiam loss\cite{chen2021exploring} functions. In this work, we use the Barlow Twins loss function on the emebddings generated from the Transformer and CGCNN encoder. As shown in Figure~\ref{fig:model}C, we initially encode both the text string representation and graph representation with their respective encoders. The MOFormer will encode the text string representation and the CGCNN will encode the graph representation. We generate an embedding of size 512 from both the encoders and use it to generate the cross-correlation matrix (Equation~\ref{eq2}). Ideally, we want cross-correlation as close to the identity matrix as both the representations generated from MOForemer and CGCNN are essentially capturing the same MOF. The Barlow Twins loss function, which we used for SSL pre-training (Equation~\ref{eq1}), tries to force the cross-correlation matrix to the identity matrix. 
\begin{equation}
    C_{ij} \overset{\Delta}{=} \frac{\sum_b \pmb{Z}_{b,i}^A \pmb{Z}_{b,j}^B}{\sqrt{(\pmb{Z}_{b,i}^A)^2}{\sqrt{(\pmb{Z}_{b,j}^B)^2}}} 
    \label{eq2}
\end{equation}
where $b$ is the batch index and $i,j$ index the 512 dimensional output from the projector ($Z^A$ and $Z^B$), $A$ is the graph representation and $B$ is text representation for the same MOF.

\begin{equation}
    L_\mathrm{{BT}} \overset{\Delta}{=} \sum_i(1-C_{ii})^2 + \lambda\sum_i\sum_{j\neq i} C_{ij}^2,
    \label{eq1}
\end{equation}
where $\pmb{C}$  is the cross-correlation matrix of embeddings from the MOFormer and CGCNN, the cross correlation matrix is given by Equation ~\ref{eq2}. The $\lambda$ used in this work is set to $0.0051$. 

Finally after pretraining the models using SSL, the encoder weights are shared during the finetuning stage (Figure~S1). The pretraining hyperparamter details are shown in Table S3 For finetuning, we initialize the model with pretrained weights and train the model for 200 epochs to generate the final prediction(Hyperparameters: Table S1 and Table S2).   The MOFormer and CGCNN models are finetuned separately. We observed that using SSL pre-training framework improved the results of both CGCNN and MOFormer consistently for all the datasets.  

\subsection{Datasets and other featurizations}
Three public MOF datasets including the CORE MOF 2019\cite{chung2019advances}, the hypothetical MOFs\cite{wilmer2012large} (hMOF), and the Boyd\&Woo\cite{boyd2019data} are combined to create a large dataset for the self-supervised pretraining. The pretraining dataset only includes MOFs with both 3D structure and MOFid available. After the removal of duplicated MOFs, the final pretraining dataset has 413535 unique MOFs. In the downstream prediction task, the MOFormer and the CGCNN are trained on the quantum MOF\cite{rosen2021machine} (QMOF) and hMOF in a supervised manner. The QMOF dataset contains 20375 MOFs each with a label of DFT calculated band gap in eV. Only 7466 MOFs in the QMOF dataset have MOFid available. On the other hand, the hMOF has 137652 MOFs, in which 102858 have available MOFid. The models are trained on hMOF with the label of \ce{CO_2} and \ce{CH_4} adsorption in mol$\cdot$kg$^{-1}$ at 0.05, 0.5, and 2.5 bar of pressure. The benchmark datasets are split into training, test, and validation sets with a ratio of 0.7-0.15-0.15. During the training, the model with the best validation performance is recorded and then tested with the test set. According to the splitting rule, MOFormer has 5226-1119-1119 QMOF data and 72000-15428-15428 hMOF data, while other models have 14262-3056-3056 QMOF data and 96356-20647-20647 hMOF data in the training, validation and test set, respectively. Although the MOFs with available MOFid form a subset of both benchmark datasets, the subset with MOFid shares the same distribution and has approximately the same mean and standard deviation compared with the original whole dataset (Figure~S2 and S3 in the supporting information). Therefore, it is fair to compare the performance of MOFormer and other models.

We also benchmarked the MOFormer and CGCNN against other non-DL based featurization methods such as the Smooth Overlap of Atomic Positions\cite{bartok2013representing, bartok2017machine, dscribe} (SOAP) and the Stoichiometric-120\cite{meredig2014combinatorial} features. SOAP is a structure-based featurization method and the Stoichiometric-120 is a structure-agnostic featurization method. The parameters used for creating SOAP features are included in the supporting information Table~S4. XGBoost\cite{chen2016xgboost} model is used to make predictions using those handcrafted features. 

\section{Results and discussion}
\subsection{QMOF}
The first dataset we benchmark models on is the QMOF dataset, where the label for each MOF is DFT calculated band gap. Lower band gap value results in better conductivity of the MOF. Accurate prediction of band gap can help to identify conductive MOFs which is useful in energy storage applications\cite{xie2020electrically, sheberla2017conductive}. The accuracy of models follows the rank of CGCNN $>$ MOFormer $>$ SOAP $>$ Stoichiometric-120 (Table.~\ref{tb:qmof}). MOFormer has 21.2$\%$ lower MAE compared with the other structure-agnostic method Stoichiometric-120. It is worth noting that structure-agnostic MOFormer outperforms structure-based SOAP with a smaller size of training set, indicating that MOFormer is capable of extracting critical features from the MOFid for energy-related property prediction. The self-supervised pretraining helps reduce the mean absolute error (MAE) of CGCNN by 6.79$\%$ and MOFormer by 5.34$\%$. The reduced error proves the improvement brought by the pretraining. 
\begin{table}[htb!]
  \centering
  \begin{adjustbox}{width=\textwidth}
  \begin{tabular}{l|lll|lll}
    \toprule
       & CGCNN\textsubscript{scratch} & CGCNN\textsubscript{pretrain} & SOAP & MOFormer\textsubscript{scratch} & MOFormer\textsubscript{pretrain} & Stoichiometric-120\\
    \midrule
    MAE (eV) & 0.275 $\pm$0.015 & \textbf{0.256$\pm$0.006} & 0.424$\pm$0.007 & 0.387$\pm$0.001 & \textbf{0.367$\pm$0.005} & 0.466$\pm$0.011 \\
    \bottomrule
  \end{tabular}
  \end{adjustbox}
  \caption{Benchmark performance of different models on the band gap prediction of the QMOF dataset. Mean absolute error (MAE, in the unit of eV) and standard deviation of 3 runs of different initial seeds of each model is reported. The left three models are structure-based and the right three models are structure-agnostic. The best performance of each category is marked as bold.}
  \label{tb:qmof}
\end{table}

\begin{figure}[htb!]
    \centering
    \includegraphics[width=\linewidth]{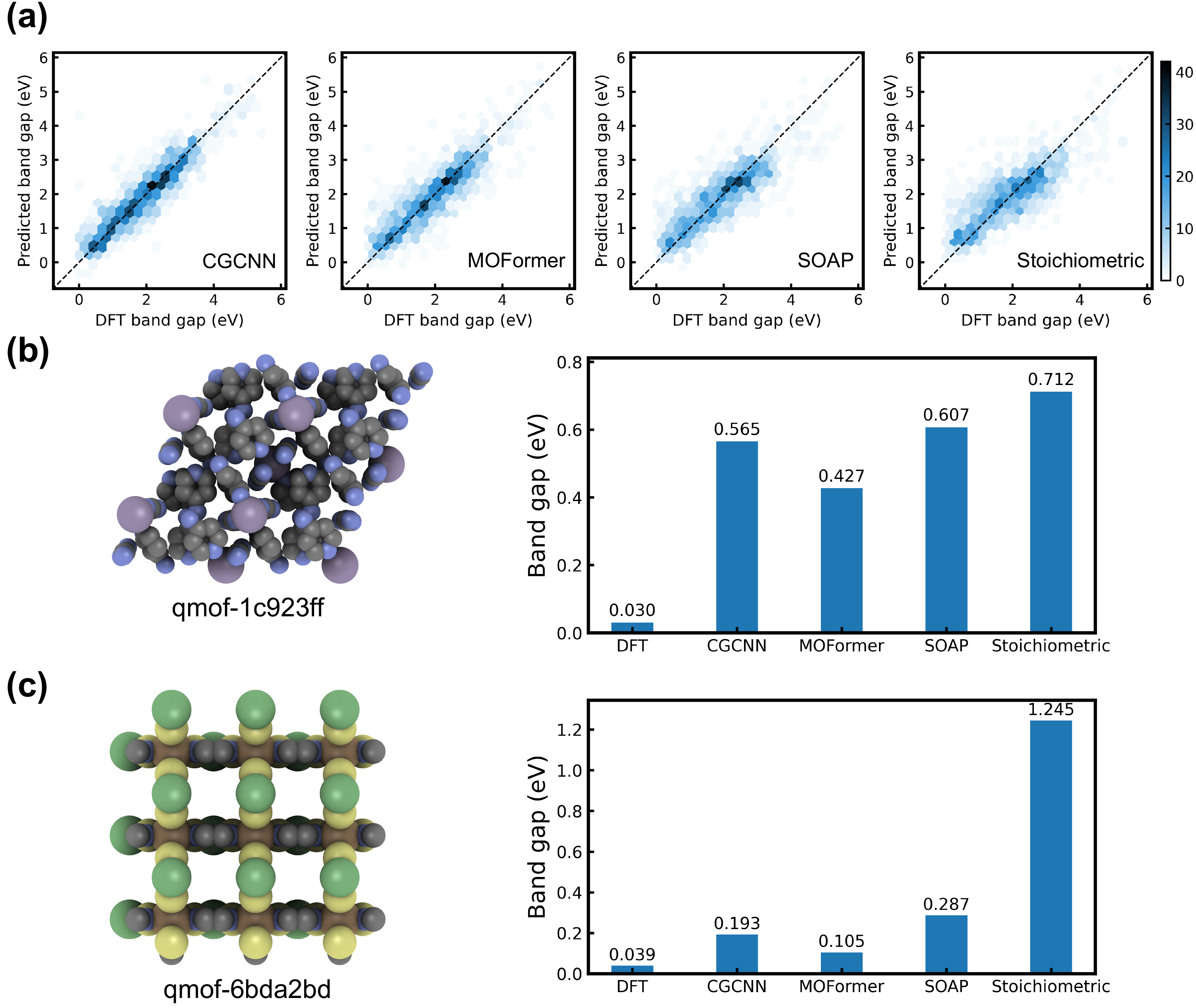}
    \caption{\textbf{(a)} The binned scatter plot shows the comparison between predicted and DFT calculated band gap for MOFs in the QMOF dataset. MOFs included in this figure are from the randomly splitted test set. Darker color of each hexagonal bin represents more data points are in the bin. Dashed line represents perfect prediction. \textbf{(b) - (c)} Visualization\cite{dubbeldam2018iraspa} of the MOF structure with lowest (qmof-1c923ff) and the second lowest (qmof-6bda2bd) band gap in the test set. The bar plot shows the comparison between predictions made by different models.}
    \label{fig:qmof}
\end{figure}
To better understand the superior performance of MOFormer and CGCNN in QMOF, we trained the 4 models with the same training set and then examined their performance on the same test set consisting 1119 randomly selected MOFs. The binned scatter plot (Figure~\ref{fig:qmof}a) shows the comparison between the predicted and the DFT calculated band gap. Darker color means more data fall in the bin. More predictions made by CGCNN and MOFormer are closer to the ground truth, especially for MOFs with band gap lower than $\leq$ 2eV. The SOAP and Stoichiometric-120 are more likely to overpredict the lower band gap. This weakness of SOAP and Stoichiometric-120 can also be confirmed by the kernel density estimation of predicted values (Figure~S4). The MOFs with the top-2 lowest band gaps in this test set are the qmof-1c923ff (0.03eV) and qmof-6bda2bd (0.039eV). Band gap prediction by MOFormer and CGCNN is much closer to the DFT calculated value than predictions by SOAP and Stoichiometric-120 (Figure~\ref{fig:qmof}b-c), especially for qmof-6bda2bd.  Accurately predicting low band gap of MOFs can lead to the discovery of conductive MOF, rendering pretrained MOFormer and CGCNN more valuable for prescreening MOFs. 

\subsection{hMOF}
The models are also benchmarked on the hMOF dataset with the label of \ce{CO_2} and \ce{CH_4} adsorption under 0.05, 0.5, and 2.5 bar of pressure. Table~\ref{tb:hmof} shows that pretrained MOFormer is constantly outperforming the other structure-agnostic Stoichiometric-120 by achieving 35$\%$-48$\%$ lower MAE. Pretrained CGCNN outperforms other models for the \ce{CO_2} adsorption prediction. The pretraining in average improves the accuracy of MOFormer by 4.3$\%$ and the CGCNN by 16.5$\%$ over all gas adsorption predictions. SOAP has surprisingly low MAE for the gas adsorption prediction, outperforming pretrained CGCNN for 2 of 3 \ce{CH_4} adsorption predictions, and the CGCNN trained from scratch for all gas adsorption predictions. The outstanding performance of SOAP on hMOF can be attributed to the low variation of elements included in the hMOF dataset. Only 11 different elements are present among all 137652 hMOFs, which is very limited compared to 79 in the QMOF dataset. Less number of elements results in much smaller and less sparse SOAP feature vector (SOAP feature has an size of 2772 for hMOFs and 19908 for QMOFs with same parameters), thus leading to the high prediction accuracy of the following XGBoost regressor. However, the high accuracy of SOAP can hardly be sustainable when it is used in exploring more diverse hypothetical MOFs. When more elements are included in the dataset, the SOAP feature vector size and sparsity increase drastically, causing the data and model to be too large to be accommodated by the memory of local machines and drop of prediction accuracy (Table~S5). MOFormer and CGCNN will not suffer from such an issue since their inputs remain invariant with increasing types of elements in the dataset, making them better choices when exploring more diverse chemical space for MOFs.

\begin{table}[htb!]
  \centering
  \begin{adjustbox}{width=\textwidth}
  \begin{tabular}{l|lll|lll}
    \toprule
      & \ce{CO_2} 0.05bar & \ce{CO_2} 0.5 bar & \ce{CO_2} 2.5 bar & \ce{CH_4} 0.05bar & \ce{CH_4} 0.5bar & \ce{CH_4} 2.5bar\\
    \midrule
    CGCNN\textsubscript{scratch} & 0.126$\pm$0.005 & 0.391$\pm$0.017 & 0.818$\pm$0.050 & 0.028$\pm$0.001 & 0.121$\pm$0.006 & 0.333$\pm$0.017 \\
    
    CGCNN\textsubscript{pretrain} & \textbf{0.110$\pm$0.001} & \textbf{0.330$\pm$0.002} & \textbf{0.645$\pm$0.003} & 0.025$\pm$0.001 & \textbf{0.099$\pm$0.001} & 0.258$\pm$0.008 \\
    
    SOAP & 0.115$\pm$0.002 & 0.339$\pm$0.004 & 0.666$\pm$0.003 & \textbf{0.022$\pm$0.001} & 0.106$\pm$0.001 & \textbf{0.239$\pm$0.002} \\
    \midrule
    MOFormer\textsubscript{scratch} & 0.178$\pm$0.002 & 0.558$\pm$0.001 & 1.000$\pm$0.013 & \textbf{0.034$\pm$0.000} & 0.174$\pm$0.002 & \textbf{0.385$\pm$0.003} \\
    
    MOFormer\textsubscript{pretrain} & \textbf{0.158$\pm$0.001} & \textbf{0.545$\pm$0.008} & \textbf{0.982$\pm$0.011} & \textbf{0.033$\pm$0.000} & \textbf{0.161$\pm$0.011} & \textbf{0.384$\pm$0.003} \\
    
    Stoichiometric-120 & 0.282$\pm$0.002 & 0.983$\pm$0.005 & 1.895$\pm$0.003 & 0.050$\pm$0.001 & 0.269$\pm$0.001 & 0.631$\pm$0.002 \\
    \bottomrule
  \end{tabular}
  \end{adjustbox}
  \caption{Benchmark performance of different models on gas adsorption prediction of the hMOF dataset. Mean absolute error (mol kg$^{-1}$) and standard deviation of 3 runs of different initial seeds of each model is reported. The top three models are structure-based and the bot three models are structure-agnostic. The best performance of each category is marked as bold.}
  \label{tb:hmof}
\end{table}

\begin{figure}[htb!]
    \centering
    \includegraphics[width=\linewidth]{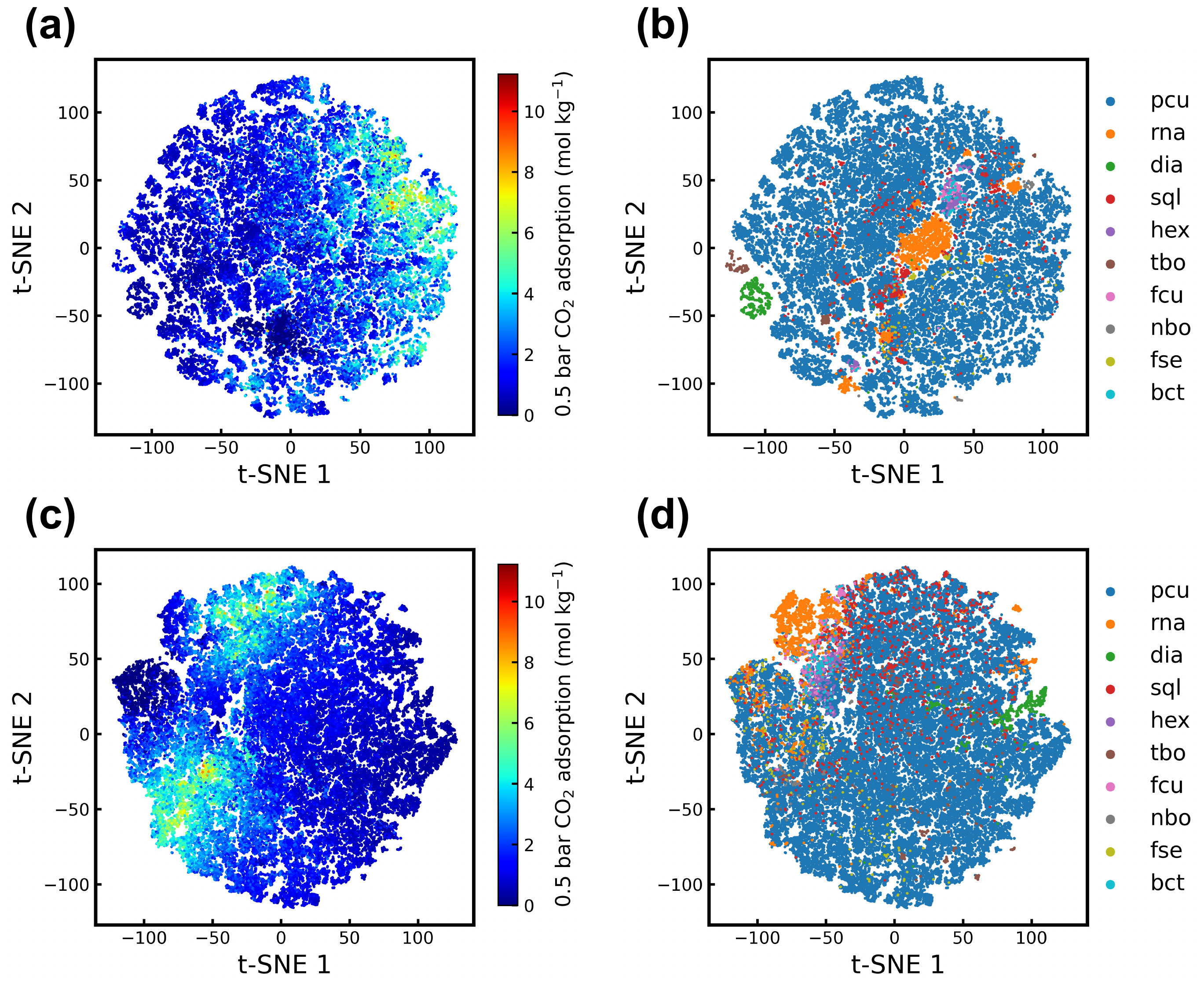}
    \caption{The t-SNE\cite{van2008visualizing} dimension reduced visualization of MOF representations learned by \textbf{(a)-(b)} the MOFormer and \textbf{(c)-(d)} the CGCNN. Each data point in \textbf{(a)} and \textbf{(c)} is colored by its \ce{CO_2} adsorption at 0.5 bar pressure, and in \textbf{(b)} and \textbf{(d)} is colored by its topology. Only MOF which has the top-10 most common topologies in hMOF dataset is shown.}
    \label{fig:tsne}
\end{figure}

The representations of MOFs learned by the MOFormer and CGCNN after finetuning are visualized to provide interpretablity to the models (Figure~\ref{fig:tsne}). Each representation is projected to the 2D space using the dimension reduction tool t-SNE\cite{van2008visualizing}. t-SNE clusters more similar data points together while placing less similar data points further away. Only MOFs which has the top-10 most common topologies in hMOF is included in Figure~\ref{fig:tsne} because they take $>99.7\%$ of the whole dataset. We can observe that CGCNN representations cluster MOF with high \ce{CO_2} adsorption more closely than MOFormer representations by comparing Figure~\ref{fig:tsne}a and \ref{fig:tsne}c. This contributes to higher prediction accuracy of CGCNN. On the other hand, MOFormer representation clusters MOFs with the same topology closer than CGCNN representation does. For example, the MOFs with dia (green) and tbo (brown) topologies form two clusters in the lower left corner of MOFormer representation visualization (Figure~\ref{fig:tsne}b). Those MOFs are much loosely clustered in the CGCNN representation visualization (Figure~\ref{fig:tsne}d). The MOFormer representation focusing on topology can be caused by the fact that gas adsorption is more dependent on the 3D structure of the MOF compared with its atom composition. The only structure-related information contained in the MOFid is the topology encoding. Therefore, more weights are on the topology after MOFormer is finetuned to predict the gas adsorption of MOFs. The input of CGCNN is the 3D structure of MOFs with atomic resolution, thus causing CGCNN to rely less on the topology for gas adsorption prediction. 

\subsection{Visualization of attention weights}
Figure 4 demonstrates the attention maps between tokens of a MOFid (qmof-ba40858) in the last MOFormer layer after finetuning on band gap prediction. The attention map can serve as a visual interpretation of how the MOFormer learns MOF representations\cite{vig2019multiscale}. We observe a strong attention in head 5 from all tokens to the metal node Ytterium, and to the topology encoding \texttt{pcu} in head 1 and 3. The attention from metal node to the topology encoding is especially high in head 1. Moreover, large attention weight can be observed between tokens in the SMILES of the SBUs in head 6. Head 1, 3, 5, and 8 also show large attention weights on the carbon and the oxygen atom, and the double bond in the organic building block. The attention weight visualization shows that MOFormer learns a representation that emphasizes on the contextual information between key components in the MOFid including important atoms (e.g. Y, O, and C) and the topology, thus leading to more accurate prediction.

\begin{figure}[htb!]
    \centering
    \includegraphics[width=\linewidth]{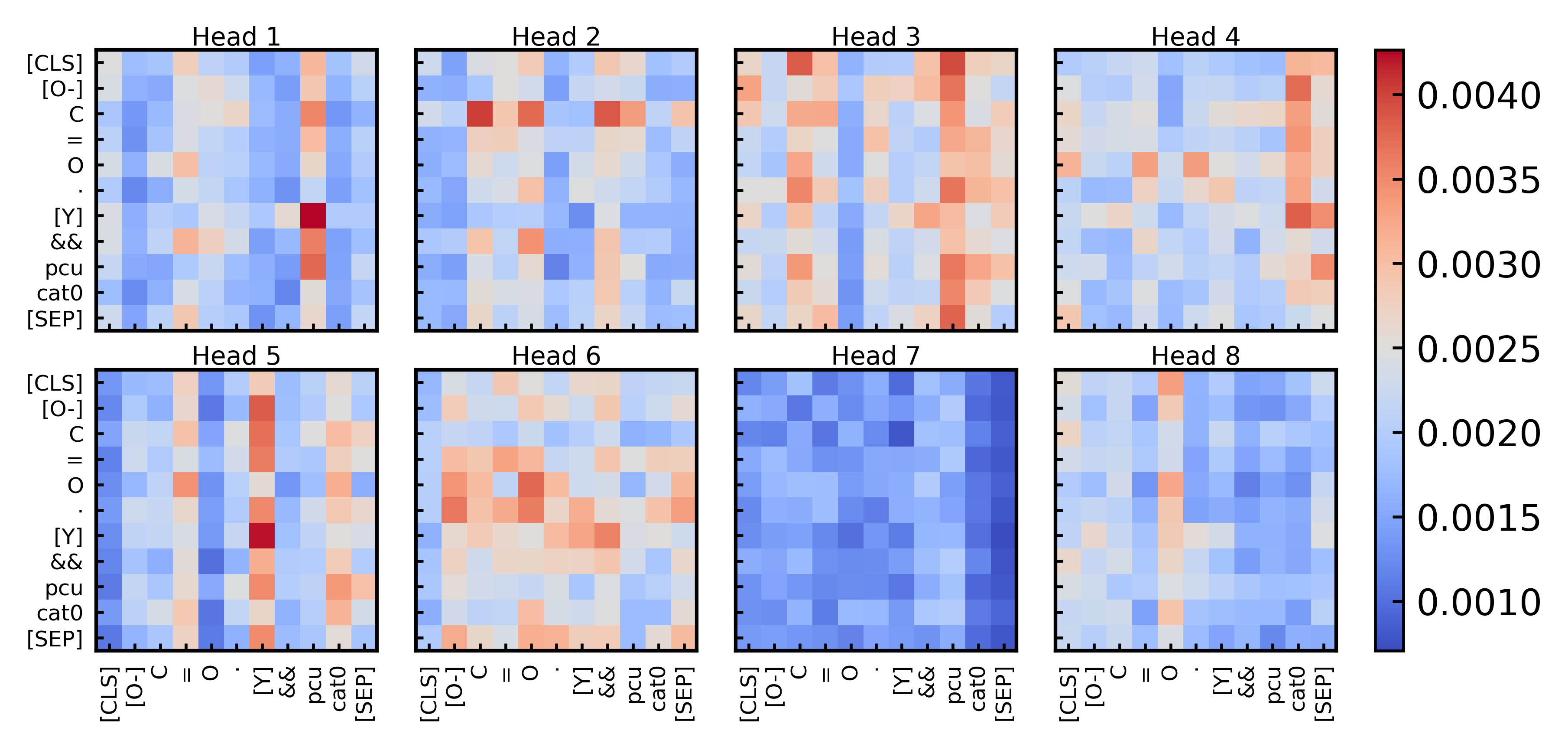}
    \caption{Heatmap of the attention between tokens (MOFid of qmof-ba40858) in different heads of the last MOFormer layer. If we index each block in the heatmap as $\text{block}_{n, i, j}$, where $i$, $j$, and $n$ is the row, column, and head index, respectively. The value of $\text{block}_{n, i, j}$ represents the attention on the $j$-th token from the $i$-th token in $n$-th heatmap.}
    \label{fig:attn_map}
\end{figure}

\subsection{Data efficiency comparison}
\begin{figure}[htb!]
    \centering
    \includegraphics[width=\linewidth]{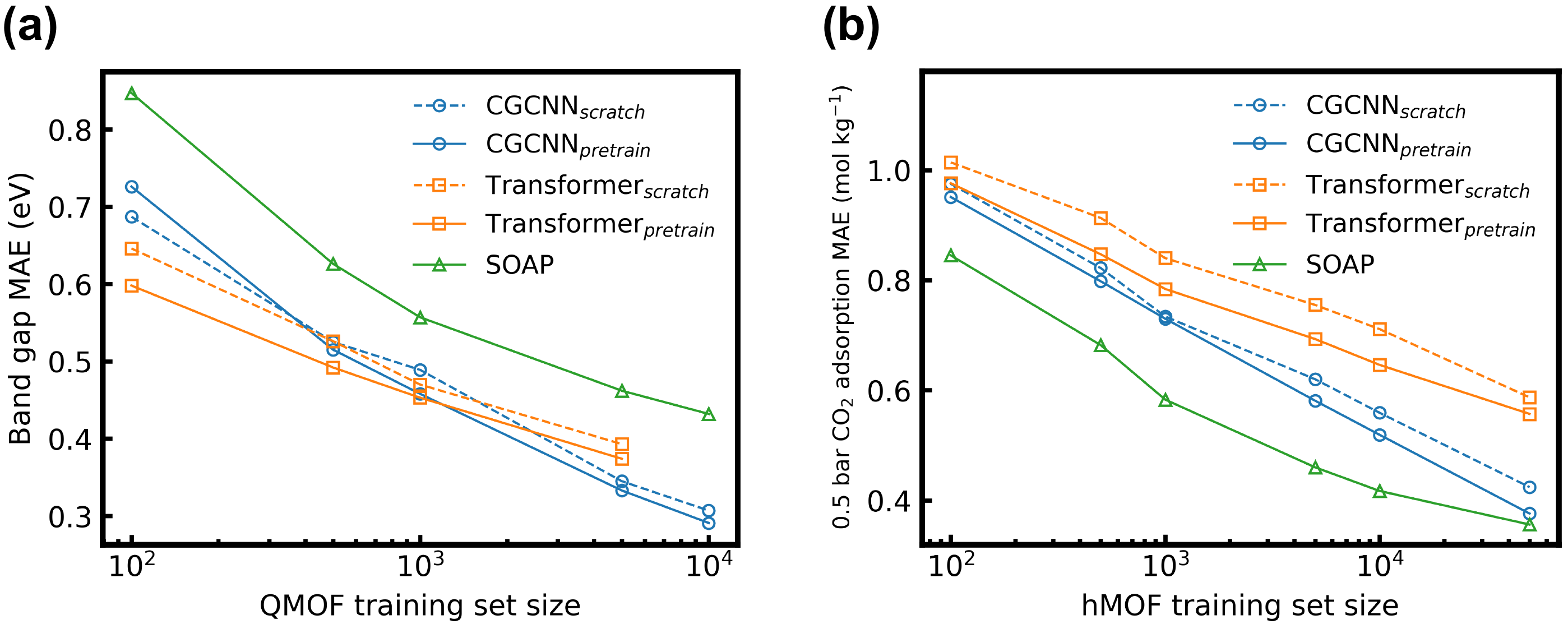}
    \caption{Data efficiency comparison between different models on the \textbf{(a)} QMOF and the \textbf{(b)} hMOF dataset. The models are trained on a subset of the training set, while the validation and test set are kept the same. The training set sizes are 100, 500, 1000, 5000, 10000, and 50000 (hMOF only). Since only less than 7500 MOFs in QMOF have available MOFid, the maximum training subset size for MOFormer on QMOF is 5000.}
    \label{fig:subset}
\end{figure}

Obtaining high-quality MOF data using experimental or DFT method can be time-consuming and expensive. A model with high data efficiency is ideal when the training data size is limited. We compared the data efficiency of different models on the QMOF and hMOF dataset (\ce{CO_2} adsorption at 0.5 bar pressure). For band gap prediction (Figure~\ref{fig:subset}a), the pretrained MOFormer outperforms CGCNN when the training set size $\leq1000$. This makes MOFormer more valuable in predicting quantum-chemical properties when the training dataset is difficult to build (i.e. experimentally synthesized MOFs). Both MOFormer and CGCNN achieve higher accuracy than SOAP regardless of the training set size on the QMOF dataset. For \ce{CO_2} adsorption prediction (Figure~\ref{fig:subset}b), CGCNN constantly achieves higher accuracy than MOFormer regardless of the training set size, indicating its higher data efficiency. CGCNN outperforms MOFormer on hMOF because the \ce{CO_2} adsorption correlates more with the MOF structure and the input to CGCNN provides more structural information than MOFid. SOAP achieves higher data efficiency than CGCNN and MOFormer on hMOF, but is eventually caught up by CGCNN after the training set size exceeds 50000. Both Figure~\ref{fig:subset}a and \ref{fig:subset}b show that pretraining consistently improves the data efficiency of MOFormer and CGCNN. Moreover, SOAP is shown to have diminishing improvement with increasing training set size, but CGCNN and MOFormer do not suffer from such an issue.

\section{Conclusion}
In summary, we propose a Transformer-based model, named as MOFormer, for structure-agnostic MOF property prediction. Taking only MOFid as input, the MOFormer model is expected to expedite the exploration of hypothetical MOFs. We also introduce a self-supervised learning framework to jointly pretrain the MOFormer and CGCNN model on large unlabeled MOF dataset to enhance their prediction accuracy in downstream tasks. Compared with another structure-agnostic method Stoichiometric-120, MOFormer achieves $21.4\%$ higher accuracy on band gap prediction, and $35\%$-$48\%$ higher accuracy on various gas adsorption prediction tasks. MOFormer even outperforms structure-based SOAP method in band gap prediction with less training data. The pretraining is further shown to improve the accuracy of MOFormer by $5.34\%$ and $4.3\%$ in average and CGCNN by $6.79\%$ and $16.5\%$ in average, for band gap and gas adsorption prediction, respectively. MOFormer and CGCNN are shown to be less likely to overpredict the band gap of MOFs compared with SOAP and Stoichiometric-120, making them better choices for prescreening conductive MOFs for energy applications. When used for gas adsorption prediction of MOFs, MOFormer relies more on the topology information compared with CGCNN because of the strong correlation between the label and the structure of MOF. Visualization of the attention weights in the last MOFormer layer reveals that the attention layers in MOFormer focus more on several important atoms and the topology to learn the representation of a MOF. Lastly, MOFormer is shown to be more data efficient than CGCNN for band gap prediction when the training set size $\leq1000$. As a structure-agnostic model, MOFormer can make rapid and accurate inference on the property of MOFs (especially for quantum-chemical properties) using arbitrarily constructed MOFid as input. Therefore, MOFormer can serve as a tool for exploring the vast chemical space of hypothetical MOFs.

\begin{acknowledgement}
This work is supported by the start-up fund from Mechanical Engineering Department at CMU.
\end{acknowledgement}

\begin{suppinfo}
Transformer and self-attention mechanism. Details of CGCNN and MOFormer model. Details of the self-supervised pretraining. Distribution of QMOF and hMOF dataset. Parameters for SOAP feature vector creation and the effect of SOAP vector length to model accuracy. Kernel density estimation of band gap prediction from different models. 

\end{suppinfo}

\section{Code Availability}
The python code of this work can be found on Github \href{https://github.com/zcao0420/MOFormer}{https://github.com/zcao0420/MOFormer}.

\newpage
\bibliography{ref}

\setcounter{table}{0}
\renewcommand{\tablename}{Table}
\setcounter{figure}{0}
\renewcommand{\figurename}{Figure}
\pagenumbering{arabic}
\renewcommand*{\thepage}{S\arabic{page}}
\maketitle
\newpage
\section{Supporting Information}
\section{Self-attention and the Transformer model}
In this work, we use the encoder part of the Transformer model as the base for the MOFormer framework. Transformer\cite{vaswani2017attention} model was proposed as a language model for sequence-to-sequence language translation. The encoder part of it learns a latent representation of the input sequence using the self-attention mechanism. Self-attention mechanism is previous used along with sequential models (such as recurrent neural network\cite{bahdanau2014neural} and LSTM\cite{hochreiter1997long}) to combat the drop in model performance with long sequence. Transformer model drop the recurrent architecture in previous sequential models, and rely solely on the self-attention mechanism to learn the representation of the input sequence. The advantage of such model architecture is threefold. First, self-attention has lower per-layer computational complexity than recurrent architectures. Second, models can be highly parallelizable with only self-attention (as self-attention can be calculated using matrix multiplication) thus can be trained much faster. Third, long-range dependencies in the input sequence can be easily learned using self-attention mechanism. 

In the self-attention mechanism, each token in the sequence is linearly projected into its corresponding query, key, and value vector ($Q$, $K$, and $V$ matrix for the entire sequence, respectively). The weights that linearly project the entire input sequence are denoted as the $W_q$, $W_k$, and $W_v$ matrices. In this work, followed by Vaswani et al.\cite{vaswani2017attention}, the scaled dot-product attention:
\begin{equation}
    Attention(Q, K, V) = \text{softmax}(\frac{QK^\top}{\sqrt{d_{emb}}})V
\end{equation}
is adopted, where $d_{emb} = 512$. Multihead attention is to have multiple linear projection matrices in parallel for the calculation of attention. The $W_q$, $W_k$, and $W_v$ in each attention head is initialized differently, thus multihead attention enables the model to learn representation of the input sequence from different subspaces. Assuming we have $h$ attention heads, the multihead attention is calculated as:
\begin{equation}
\begin{split}
    Multihead Attn(Q, K, V) &= \text{Concat}(head_1 \dots head_h)W_o\\
    \text{where}\; head_i &= Attention(Q_i, K_i, V_i)
\end{split}
\end{equation}
$W_o$ is a matrix to linearly project the concatenated attention from each head; $Q_i$, $K_i$, and $V_i$ is the query, key, and value matrix of the $i$-th head, respectively. In this work, the number of head  $h=8$. The dimension of the $W_q^i$, $W_k^i$, and $W_v^i$ (linear projection matrices in the $i$-th head) is $W_q^i \in \R ^{d_{emb}\times d_k}$, $W_k^i \in \R ^{d_{emb}\times d_k}$, $W_v^i \in \R ^{d_{emb}\times d_v}$, where $d_{emb}=512$ and $d_v=d_k=d_{emb}/h=64$. Moreover, $W_o\in \R ^{d_{emb}\times d_{emb}}$.
The Transformer encoder in this work contains 6 self-attention layers (Figure 1b in the manuscript), in which the MLP has a dimension of 512.
\newpage

\section{Training Details}
In this section, we describe in details the hyperparameters that we used for training the models. Our framework consists of MOFormer (transformer), CGGNN and the SSL pretraining done to improve the performance of both the GNN and the transformer model. During the finetuning stage, a MLP regression head is attached to the encoder part (MOFormer or CGCNN) to make prediction. The regression head has 4 layers, with 512, 256, 128 and 64 neurons in each layer, respectively.

\subsection{MOFormer}
The MOFormer\textsubscript{Finetuning} hyperparameters are shown in Table~\ref{tb:PreHyperparam}. Different learning rate (LR) are used for the encoder part (Transformer encoder) and the MLP regression head. We also train the MOFormer\textsubscript{Scratch} to evaluate the improvement in performance after SSL pre-training. 

\begin{table}[htb!]
\renewcommand{\thetable}{S\arabic{table}}
  \centering
  \resizebox{\textwidth}{!}{\begin{tabular}{l|llllllll}
    \toprule
    Hyperparameters & Batch Size & LR$_{Encoder}$ & LR$_{MLP}$ & Optimizer & Epochs & Embed. Size & Train/Val & Weight Decay \\
    \midrule
    MOFormer\textsubscript{Scratch} & 64 &  0.00005 &  0.01& Adam & 200 & 512 &  0.7/0.15/0.15  & $10^{-6}$\\
    MOFormer\textsubscript{Finetuning} & 64 &  0.00005& 0.01 & Adam & 200 & 512 &  0.7/0.15/0.15 & $10^{-6}$ \\
    \bottomrule
  \end{tabular}}
  \caption{Finetuning hyperparameters for the MOFormer model. LR indicate the learing rate}
  \label{tb:PreHyperparam}
\end{table}

\subsection{CGCNN}
The CGCNN model is structure based model that takes the ``.cif" file as input. The MOFormer is a structure agnostic method, however, information like neighbourhood of atoms and geometry of the crystal are often lost in the text representation. To ensure we also have a structure based benchmark to predict MOF properties we also evaluate the performance of the CGCNN model\cite{xie2018crystal}. The CGCNN model with its information rich features often outperforms the structure agnostic MOFormer. We also pretrain the MOFormer and the CGCNN together using the self-supervised learning. The performance of pretrained CGCNN\textsubscript{Pretrain} model is found to be higher than CGCNN\textsubscript{Scratch} model trained from Scratch without pretraining. The hyperparamters we used for the model are shown in Table~\ref{tb:PreHyperparamCG}.

\begin{table}[htb!]
\renewcommand{\thetable}{S\arabic{table}}
  \centering
  \resizebox{\textwidth}{!}{\begin{tabular}{l|llllllll}
    \toprule
    Hyperparameters & Batch Size &Number of layers & LR & Optimizer & Epochs & Embed. Size & Train/Val/Test & Weight Decay \\
    \midrule
    CGCNN\textsubscript{Scratch} & 128 & 3 &  0.01 & Adam & 200 & 512 &  0.7/0.15/0.15  & $10^{-6}$ \\
    CGCNN\textsubscript{Finetuning}& 128 & 3 &  0.002 & Adam & 200 & 512 &  0.7/0.15/0.15  &  $10^{-6}$\\
    \bottomrule
  \end{tabular}}
  \caption{Finetuning hyperparameters for the CGCNN model. LR denotes the learning rate for the models.}
  \label{tb:PreHyperparamCG}
\end{table}

\subsection{SSL - Pretraining}
\begin{figure}[htb!]
\renewcommand{\thefigure}{S\arabic{figure}}
    \centering
    \includegraphics[width=0.7\linewidth]{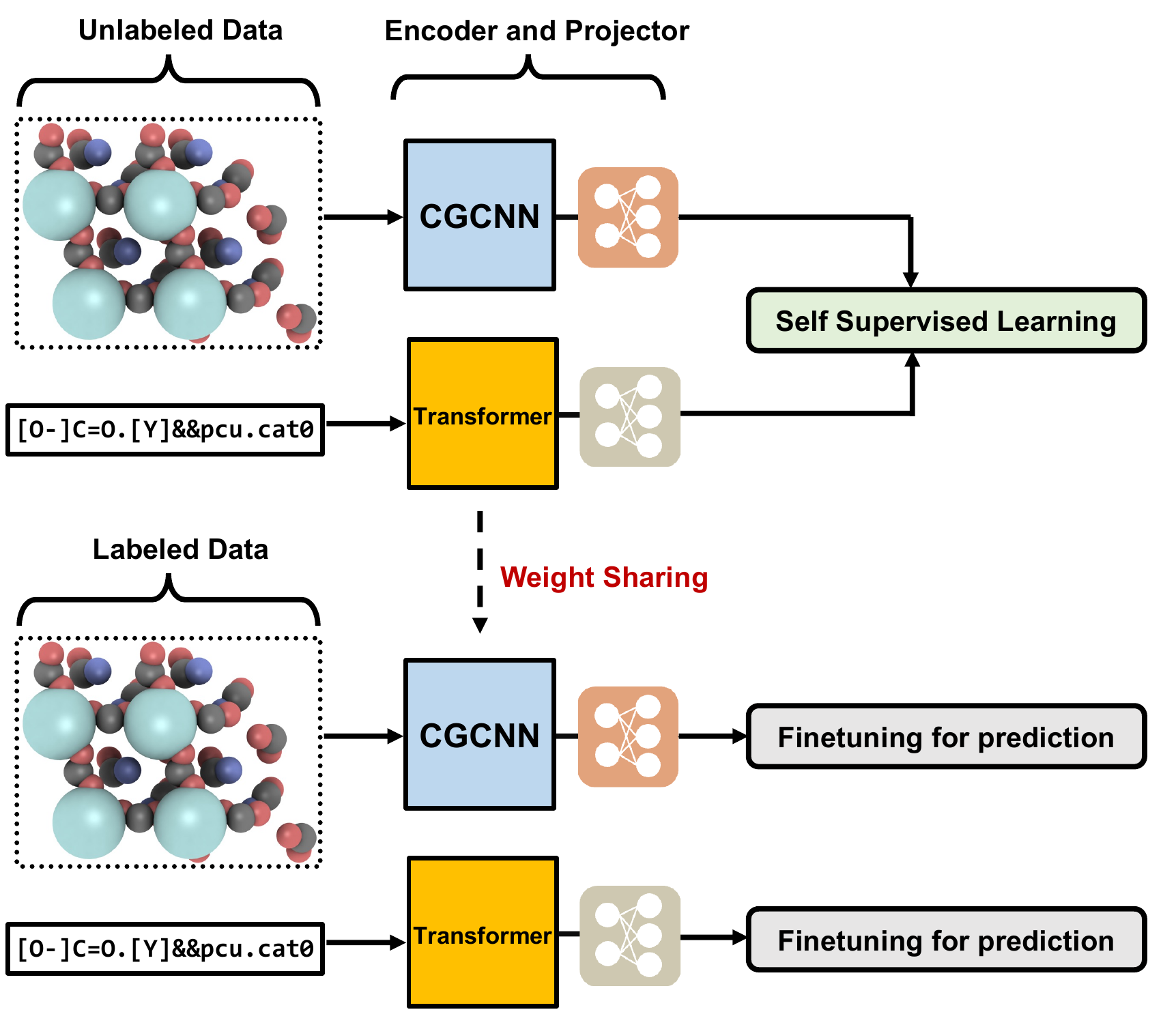}
    \caption{The SSL pretraining and finetuning framework. In this work, we use self supervised learning to pre-train both the CGCNN model and the MOFormer using self-supervised learning. The pre-trained weights are then shared during finetuning for down stream tasks. Using such a technique help us improve the perfromance of both the CGCNN and MOFormer. }
    \label{fig:ssl_mof}
\end{figure}

Structure agnostic models often lack the geometry and atom neighborhood information leading to an information bottleneck for these models. To mitigate this issue, we develop the SSL pre-training framework. In the SSL pre-training framework we have the MOFormer branch and the CGCNN branch, the goal of the SSL pre-training is to ensure that the representations from both the MOFormer and the CGCNN are similar. The pre-training helps both the MOFormer model and CGCNN in a way that it tries to leverage the embeddings learnt by the other model. The model benefit mutually from each other and the performance enhancements can be seen for both models during the finetuning stage (Figure ~\ref{fig:ssl_mof}). The hyperparamters we used during pretraining the model are shown in  Table ~\ref{tb:PreHyperparamPre}
\begin{table}[htb!]
\renewcommand{\thetable}{S\arabic{table}}
  \centering
  \resizebox{\textwidth}{!}{\begin{tabular}{l|llllllll}
    \toprule
    Hyperparameters & Batch Size & Learning Rate & Optimizer & Epochs & Embed. Size & Train/Val \\
    \midrule
    CGCNN\textsubscript{Pretrain} & 32 &  0.00001 & Adam & 15 & 512 &  0.95/0.05   \\
    MOFormer\textsubscript{Pretrain}& 32 &  0.00001 & Adam & 15 & 512 &  0.95/0.05   \\
    \bottomrule
  \end{tabular}}
  \caption{Hyperparameters for Pretraining the model.}
  \label{tb:PreHyperparamPre}
\end{table}

\newpage

\section{Distribution of the benchmark datasets}
\begin{figure}[htb!]
\renewcommand{\thefigure}{S\arabic{figure}}
    \centering
    \includegraphics[width=0.5\linewidth]{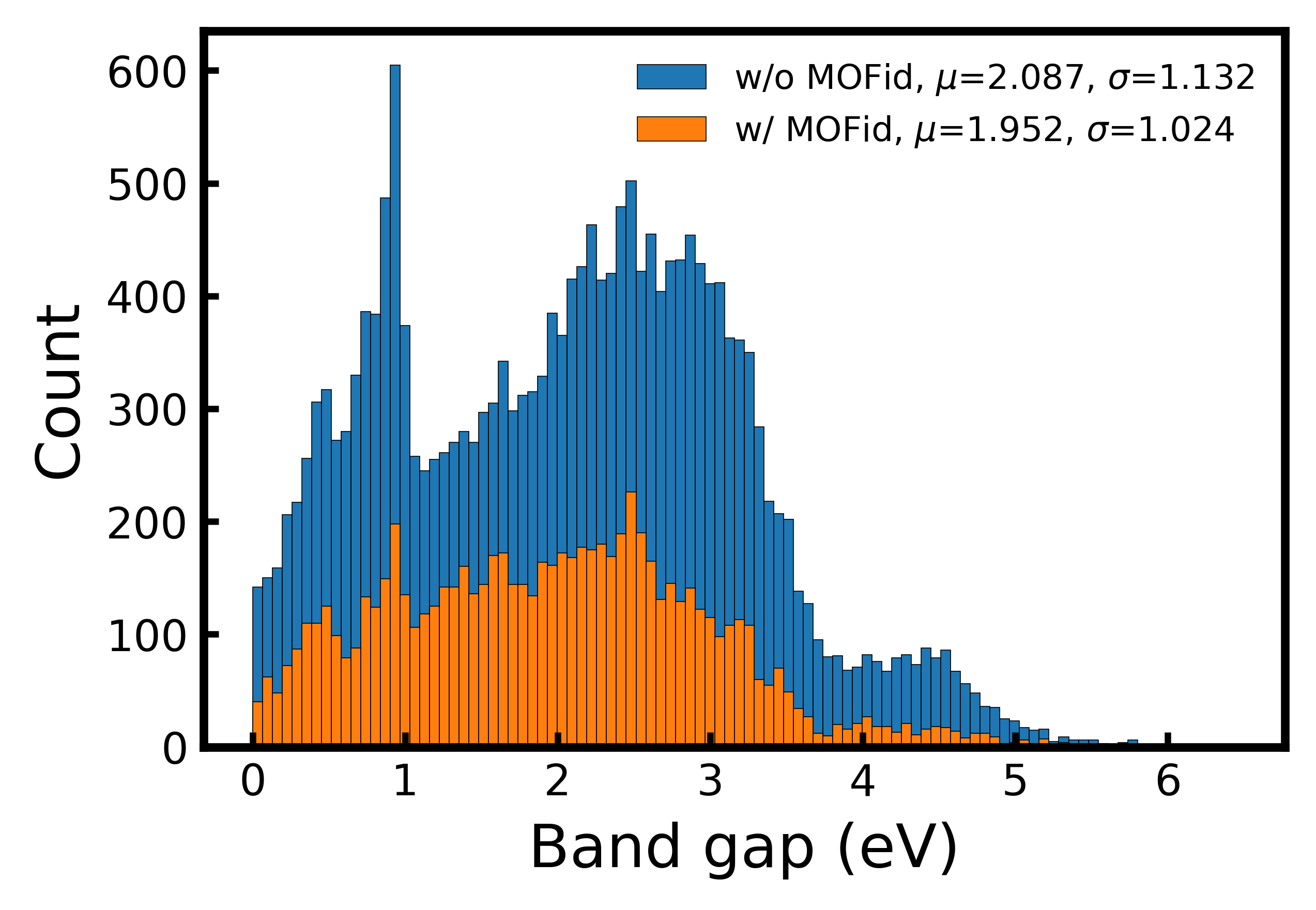}
    \caption{Histogram showing the distribution of QMOF dataset. Blue bars are all MOFs in the QMOF dataset. Orange bars are MOFs with MOFid available. Mean $\mu$ and standard deviation $\sigma$ are reported.}
    \label{fig:qmof_dist}
\end{figure}

Since only part of MOF in each QMOF\cite{rosen2021machine} and hMOF\cite{wilmer2012large} dataset have available MOFid, we plot the distribution of the whole datasets and the subsets with MOFid. Figure~\ref{fig:qmof_dist} shows that the difference between the average band gap value between the whole dataset and the subset is as small as 0.135. The distribution between the data are also similar.

\begin{figure}[htb!]
\renewcommand{\thefigure}{S\arabic{figure}}
    \centering
    \includegraphics[width=\linewidth]{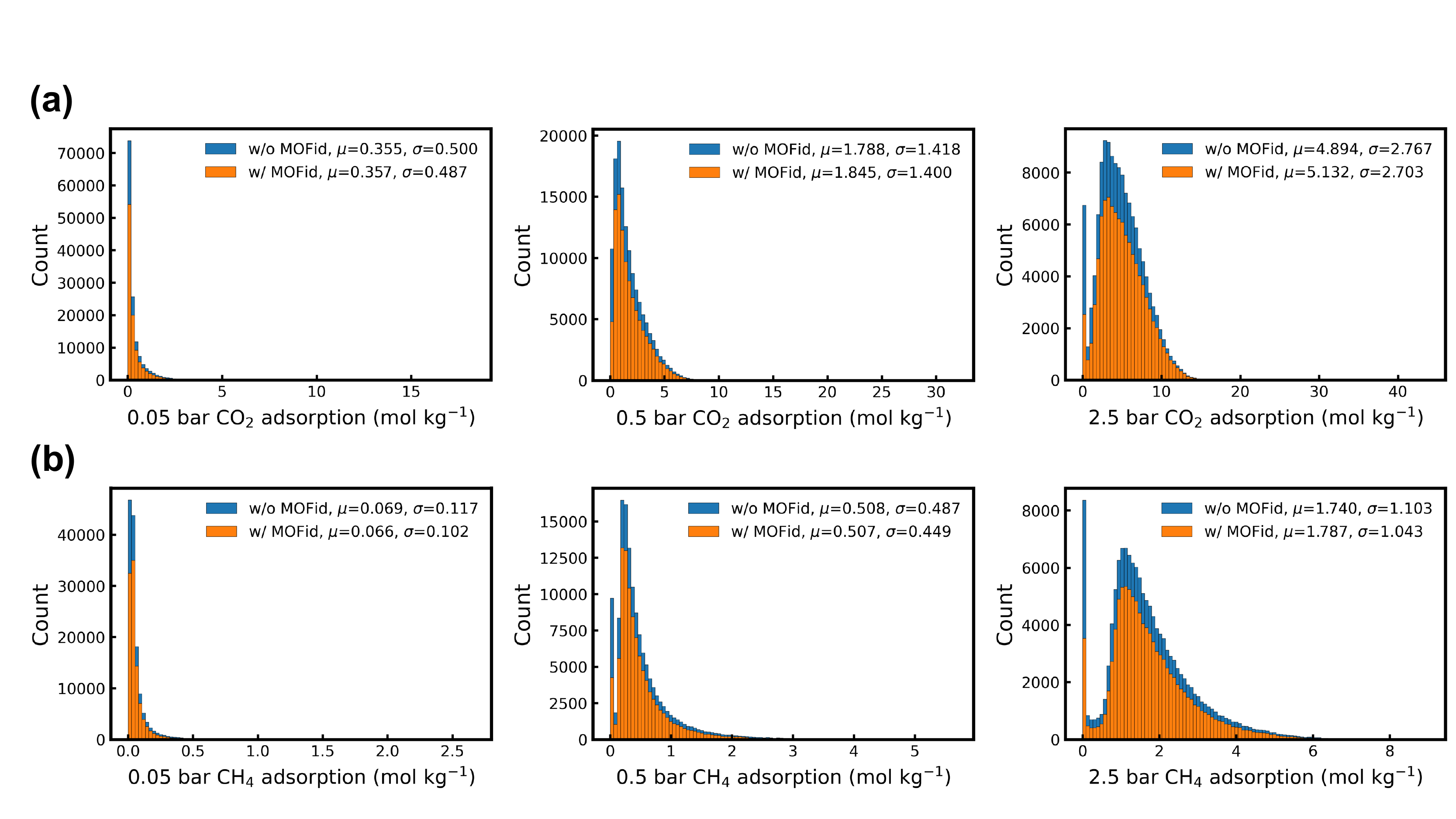}
    \caption{Histogram showing the distribution of hMOF dataset with the label of \textbf{(a)} \ce{CO_2} adsoprtion and \textbf{(b)} \ce{CH_4} adsorption at 0.05, 0.5, and 2.5 bar of pressure. Blue bars are all MOFs in the hMOF dataset. Orange bars are MOFs with MOFid available. Mean $\mu$ and standard deviation $\sigma$ are reported.}
    \label{fig:hmof_dist}
\end{figure}

Figure~\ref{fig:hmof_dist} shows that the difference between the average \ce{CO_2}/\ce{CH_4} adsorption value between the whole dataset and the subset is very small. The distribution between the data are also similar. Therefore the benchmark results comparisons between MOFormer and other models on both datasets are fair.

\newpage

\section{Parameters to create SOAP feature vector}
The SOAP\cite{bartok2013representing} feature vector is created using the DScribe package\cite{dscribe}. Table~\ref{tb:soap} records the parameters for creating SOAP feature vector, which are chosen arbitrarily based on previous works\cite{fung2021benchmarking, rosen2021machine}.

\begin{table}[htb!]
\renewcommand{\thetable}{S\arabic{table}}
  \centering
  \small
  \begin{tabular}{llllll}
    \toprule
    rcut & $\sigma$ & nmax & lmax & average method & crossover\\
    \midrule
    8& 0.2 & 8 & 6 & Inner average & False   \\
    \bottomrule
  \end{tabular}
  \caption{Parameters used for creating SOAP feature vector}
  \label{tb:soap}
\end{table}

The length of the SOAP feature vector varies greatly based on the number of different atom elements in the dataset. When the MOFs in a dataset includes a large variation of elements (such as QMOF, 79 elements), the SOAP feature vector will be very long and sparse. However, when the MOFs in the dataset consist of only a few different elements (such as hMOF, 11 elements), the SOAP feature vector will be relatively short and less sparse. For example, SOAP feature vector created using the 11 elements in hMOF has a length of 2772. The SOAP feature vector created using the 79 elements in QMOF has a length of 19908, given that nmax and lmax parameters are kept the same. The high accuracy of SOAP feature vector on hMOF is a direct result of the small variation of element type in the dataset. Assuming that we create the SOAP feature vectors for hMOFs using double the original number of elements (22 instead of the original 11), the length of SOAP feature vector will be doubled to 5544. The performance of XGBoost using the new SOAP feature is significantly reduced (Table~\ref{tb:soap2}). Therefore, we can conclude that SOAP feature is not a suitable choice when exploring the chemical space for MOFs where elements are diverse.

\begin{table}[htb!]
\renewcommand{\thetable}{S\arabic{table}}
  \centering
  \small
  \begin{tabular}{llll}
    \toprule
    Model & $\#$ elements & SOAP vector length & hMOF 2.5 bar \ce{CH_4} adsorption \\
    \midrule
    SOAP& 11 & 2772 & 0.239$\pm$0.002  \\
    SOAP& 22 & 5544& 0.288$\pm$0.002   \\
    CGCNN\textsubscript{pretrain}& - & - & 0.258$\pm$0.008   \\
    \bottomrule
  \end{tabular}
  \caption{Parameters used for creating SOAP feature vector}
  \label{tb:soap2}
\end{table}
\newpage
\section{Comparison between models' performance in QMOF dataset}
\begin{figure}[htb!]
\renewcommand{\thefigure}{S\arabic{figure}}
    \centering
    \includegraphics[width=\linewidth]{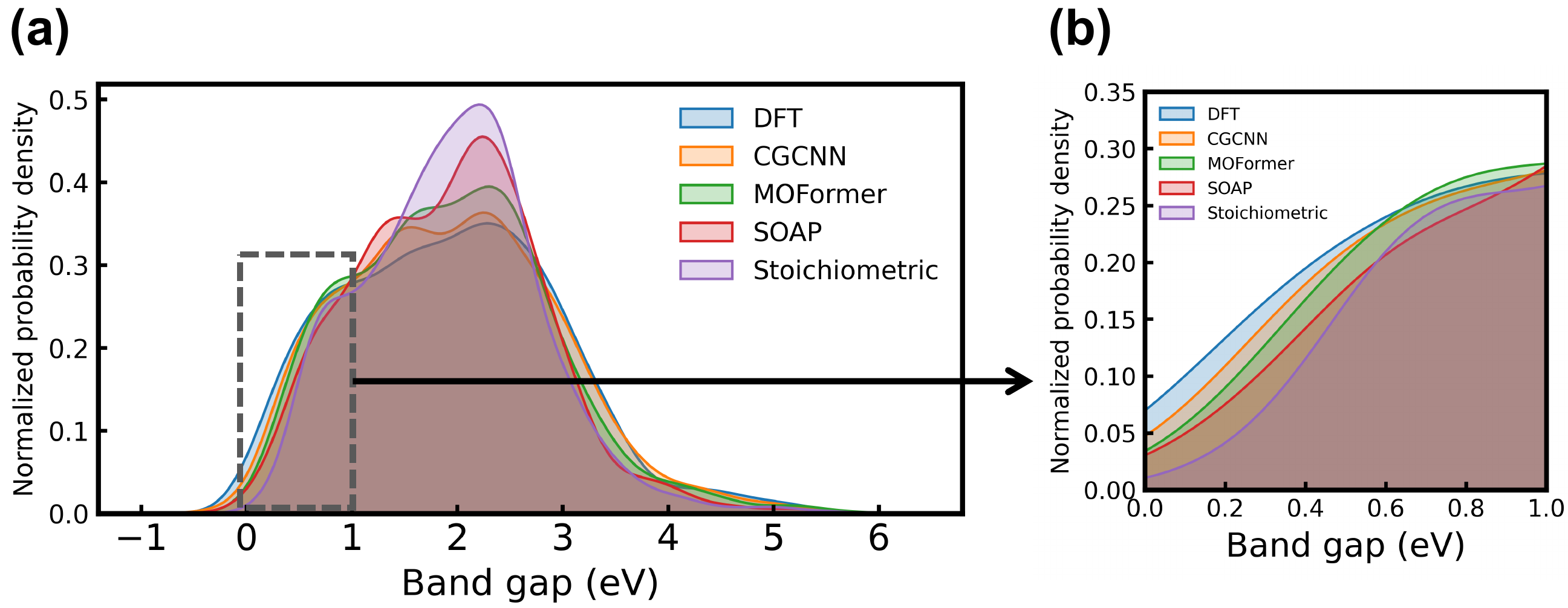}
    \caption{\textbf{(a)} The kernel density estimation (KDE) plot of the predicted band gap values of MOFs from different models. \textbf{(b)} Zoom-in view of the distribution of MOFs with band gap $\leq$ 1eV}
    \label{fig:qmof_kde}
\end{figure}

Figure~\ref{fig:qmof_kde} shows the kernel density estimation (KDE) plot of the predicted band gap values of MOFs from different models. The MOFs are in a randomly selected test set from the QMOF dataset. The ground truth is the DFT calculated band gap, which is shown by the blue curve. SOAP and Stoichiometric-120 features with XGBoost model underpredict low bandgap values more frequently than CGCNN and MOFormer.
\newpage

\newpage

\end{document}